\documentclass[11pt,a4paper]{article}
\usepackage[hyperref]{emnlp-ijcnlp-2019}
\usepackage{times}
\usepackage{latexsym}

\usepackage{url}

\usepackage{multirow}
\usepackage{graphicx}
\usepackage{CJKutf8}
\usepackage{enumerate}

\aclfinalcopy 

\title{A Pilot Study for Chinese SQL Semantic Parsing}

\author{
	Qingkai Min , Yuefeng Shi  \and Yue Zhang \\ 
	School of Engineering, Westlake University, China \\
	Institute of Advanced Technology, Westlake Institute for Advanced Study \\
	{\tt \{minqingkai, shiyuefeng, zhangyue\}@westlake.edu.cn} \\
}

\date{}

\begin{document}

\begin{CJK}{UTF8}{gbsn}

\maketitle
\begin{abstract}
  The task of semantic parsing is highly useful for dialogue and question answering systems. Many datasets have been proposed to map natural language text into SQL, among which the recent Spider dataset provides cross-domain samples with multiple tables and complex queries. We build a Spider dataset for Chinese, which is currently a low-resource language in this task area. Interesting research questions arise from the uniqueness of the language, which requires word segmentation, and also from the fact that SQL keywords and columns of DB tables are typically written in English. We compare character- and word-based encoders for a semantic parser, and different embedding schemes. Results show that word-based semantic parser is subject to segmentation errors and cross-lingual word embeddings are useful for text-to-SQL.
\end{abstract}

\section{Introduction}

The task of semantic parsing is highly useful for tasks such as dialogue \cite{chen2013unsupervised, gupta2018semantic, einolghozati2019improving} and question answering \cite{gildea2002automatic,yih2015semantic, reddy2016transforming}. Among a wide range of possible semantic representations, SQL offers a standardized interface to knowledge bases across tasks \cite{astrova2009rules, xu2017sqlnet,dong2018coarse, lee2011ysmart}. Recently, \citet{yu2018spider} released a manually labelled dataset for parsing natural language questions into complex SQL, which facilitates related research.

\citet{yu2018spider}'s dataset is exclusive for English questions. Intuitively, the same semantic parsing task can be applied cross-lingual, since SQL is a universal semantic representation and database interface. However, for languages other than English, there can be added difficulties parsing into SQL. Take Chinese for example, the additional challenges can be at least two-fold. First, structures of relational databases, in particular names and column names of DB tables, are typically represented in English. This adds to the challenges to question-to-DB mapping. Second, the basic semantic unit for denoting columns or cells can be words, but word segmentation can be erroneous. It is also interesting to study the influence of other linguistic characteristics of Chinese, such as zero-pronoun, on its SQL parsing.

We investigate parsing Chinese questions to SQL by creating a first dataset, and empirically evaluating a strong baseline model on the dataset. In particular, we translate the Spider \cite{yu2018spider} dataset into Chinese. Using the model of \citet{yu2018syntaxsqlnet}, we compare several key model configurations.

Results show that our human-translated dataset is significantly more reliable compared to a dataset composed of machine-translated questions. In addition, the overall accuracy for Chinese SQL semantic parsing can be comparable to that for English. We found that cross-lingual word embeddings are useful for matching Chinese questions with English table columns and keywords and that language characteristics have a significant influence on parsing results. We release our dataset named CSpider and code at https://github.com/taolusi/chisp.

\section{Related Work}

Existing datasets for semantic parsing can be classified into two major categories. The first uses logic for semantic representation, including ATIS \cite{price1990evaluation, dahl1994expanding} and GeroQuery \cite{zelle1996learning}. The second and dominant category of datasets uses SQL, which includes Restaurants \cite{tang2001using,popescu2003towards}, Academic \cite{iyer2017learning}, Yelp and IMDB \cite{yaghmazadeh2017sqlizer}, Advising \cite{finegan2018improving} and the recently proposed WikiSQL \cite{zhong2017seq2sql} and Spider \cite{yu2018spider}. One salient difference between Spider and prior work is that Spider uses different databases across domains for training and testing, which can verify the generalization power of a semantic parsing model. Compared with WikiSQL, Spider further has multiple tables in each database and correspondingly more complex queries. We thus consider Spider for sourcing our dataset. Existing semantic parsing datasets for Chinese include a small corpus for assigning semantic roles \cite{sun2004shallow} and SemEval-2016 Task 9 for Chinese semantic dependency parsing \cite{che2012semeval}, but these data are not related to SQL. To our knowledge, we are the first to release a Chinese SQL semantic parsing dataset. 

There has been a line of work improving the model of \citet{yu2018syntaxsqlnet} since the release of the Spider dataset \cite{guo-etal-2019-towards,bogin-etal-2019-representing,lin2019grammar}. At the time of our investigation, however, the models are not published. We thus chose the model of \citet{yu2018syntaxsqlnet} as our baseline. The choice of more different neural models is orthogonal to our dataset contribution, but can empirically give more insights about the conclusions.

\section{Dataset}

We translate all English questions in the Spider dataset into Chinese.\footnote{Note that we do not translate the database schema (i.e., column names) into Chinese because in practice databases have English schema and Chinese contents in the industry.} The work is undertaken by 2 NLP researchers and 1 computer science student. Each question is first translated by one annotator, and then cross-checked and corrected by a second annotator. Finally, a third annotator verifies the original and corrected versions. Statistics of the dataset are shown in Table \ref{tab:statistics}. There are originally 10181 questions from Spider, but only 9691 for the training and development sets are publicly available. We thus translated these sentences only. Following the database split setting of \newcite{yu2018spider}, we make training, development and test sets split in a way that no database overlaps in them as shown in Table \ref{tab:statistics}.

\begin{table}[t]
\resizebox{0.475\textwidth}{!}{%
\begin{tabular}{llcccc}
\hline
 &  & \# Q & \# SQL & \# DB & \# Table/DB \\ \hline
English & all & 10181 & 5693 & 200 & 5.1 \\ \hline
\multirow{4}{*}{Chinese} & all & 9691 & 5263 & 166 & 5.28 \\
 & train & 6831 & 3493 & 99 & 5.38 \\
 & dev & 954 & 589 & 25 & 4.16 \\
 & test & 1906 & 1193 & 42 & 5.69 \\ \hline
\end{tabular}%
}
\caption{Comparisons between Spider and Chinese Spider datasets.}
\label{tab:statistics}
\end{table}

\begin{figure}[t]
	\centering 
	\includegraphics[width=1\linewidth]{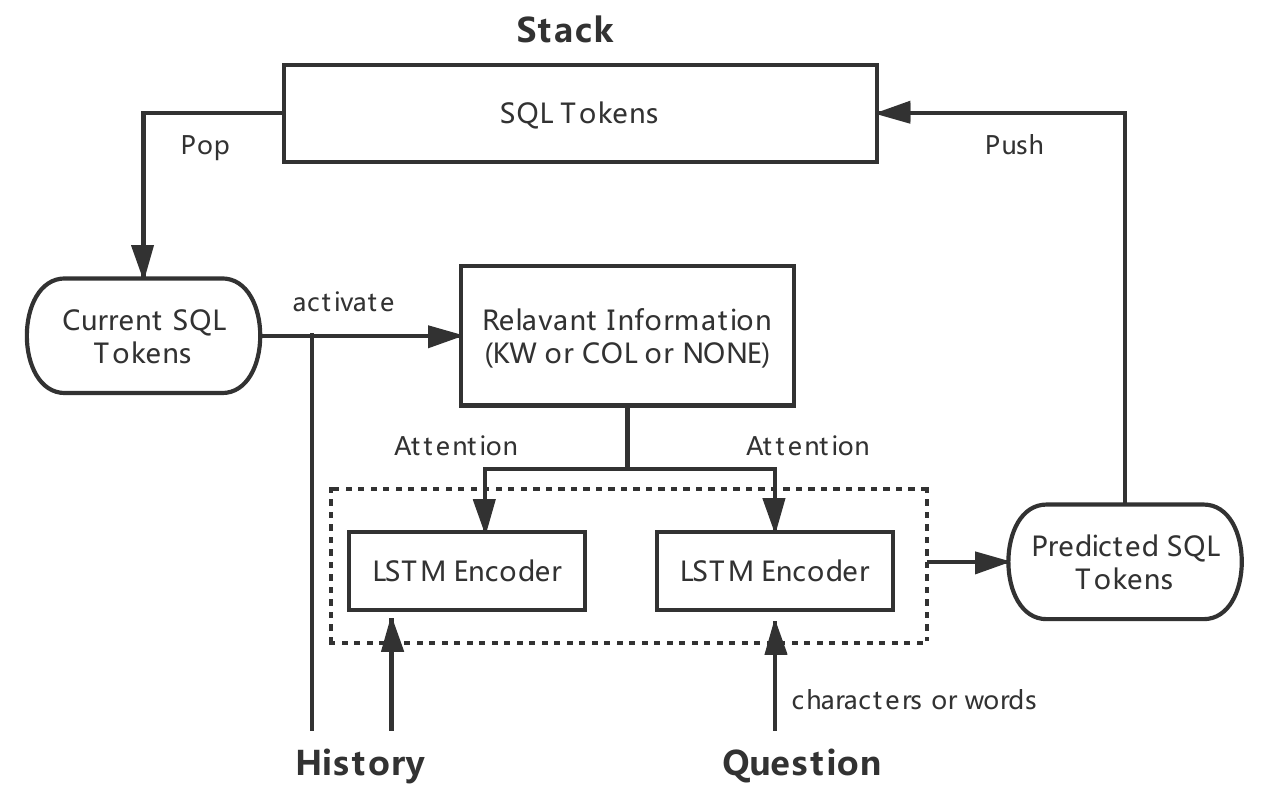}
	\caption{Overall structure of the Model.}
	\label{fig:model}
\end{figure}

The translation work is performed on a database to database basis. For each database, the same translator translates relevant inquiries sentence by sentence. The translator is asked to read the original question as well as the SQL query before making its Chinese translation. If the literal translation is possible, the translator is asked to stick to the original sentence style as much as feasible. For complex questions, the translator is allowed to rephrase the English question, so that the most natural Chinese translation is made. In addition, we keep the diversity of style in the English dataset by matching different English expressions to different Chinese expressions. A sample of our dataset is shown in Table \ref{tab:examples}. Our dataset is named CSpider.

\begin{table}[t]
\centering
\resizebox{0.475\textwidth}{!}{%
\begin{tabular}{l}
\hline
\textbf{Sample 1: applying only one table in one database.} \\ \hline
\textbf{SQL Query} \\
SELECT area FROM state WHERE state\_name = "New Mexico"; \\
\textbf{English Question} \\
What is the size of New Mexico? \\
\textbf{Translated Chinese Question} \\
新墨西哥州的面积是多少？ \\ \hline
\textbf{Sample 2: applying multiple tables in one database.} \\ \hline
\textbf{SQL Query} \\
\begin{tabular}[c]{@{}l@{}}SELECT T2.star\_rating\_description FROM HOTELS AS T1 JOIN \\ Ref\_Hotel\_Star\_Ratings AS T2 ON T1.star\_rating\_code = T2.star\_rating\_code \\ 
WHERE T1.price\_range  \textgreater  10000;\end{tabular} \\
\textbf{English Question} \\
Give me the star rating descriptions of the hotels that cost more than 10000. \\
\textbf{Translated Chinese Question} \\
给出费用超过10000的酒店星级的描述。 \\ \hline
\textbf{Sample 3:with a nested SQL query.} \\ \hline
\textbf{SQL Query} \\
\begin{tabular}[c]{@{}l@{}}SELECT T1.staff\_name ,  T1.staff\_id FROM Staff AS T1 JOIN  Fault\_Log \\
AS T2 ON T1.staff\_id  =  T2.recorded\_by\_staff\_id EXCEPT SELECT \\
T3.staff\_name, T3.staff\_id FROM Staff AS T3 JOIN Engineer\_Visits AS \\
T4 ON T3.staff\_id  =  T4.contact\_staff\_id";\end{tabular} \\
\textbf{English Question} \\
\begin{tabular}[c]{@{}l@{}}What is the name and ID of the staff who recorded the fault log but has not \\
contacted any visiting engineers?\end{tabular} \\
\textbf{Translated Chinese Question} \\
那些记录了错误报告但没有联系任何到访工程师的职工的姓名和ID \\
是什么？ \\ \hline
\end{tabular}%
}
\caption{Example questions corresponding to SQL.}
\label{tab:examples}
\end{table}

\section{Model}

We use the neural semantic parsing method of \citet{yu2018syntaxsqlnet} as the baseline model, which can be regarded as a sequence-to-tree model. In particular, the input question is encoded using an LSTM sequence encoder, and the output is a SQL query in its syntactic tree form. The tree is generated incrementally top-down, in a pre-order traversal sequence. Tree nodes include keyword nodes (e.g., SELECT, WHERE, EXCEPT) and table column name nodes (e.g., ID, City, Surname, which are defined in specific tables), which are represented in respective embedding spaces. Each keyword or column is generated by attention to the embedding space using the question representation as a key. A stack is used for incremental decoding, where the whole output history is leveraged as a feature for deciding the next term. This method gives the current released state-of-the-art results while submitting this paper. We provide a visualization of the model in Figure \ref{fig:model}.

\section{Experiments}
We focus on comparing different word segmentation methods and different embedding representations. As discussed above, column names are selected by attention over column embeddings using sentence representation as a key. Hence there must be a link between the embeddings of columns and those of the questions. Since columns are written in English and questions in Chinese, we consider two embedding methods. The first method is to use two separate sets of embeddings for Chinese and English, respectively. We use Glove \cite{pennington2014glove}\footnote{https://nlp.stanford.edu/projects/glove/} for embeddings of English keywords, column names etc., and Tencent  embeddings \cite{song2018directional}\footnote{https://ai.tencent.com/ailab/nlp/embedding.html\label{tencent}} for Chinese. The second method is to directly use the cross-lingual word embeddings. To this end, the Tencent multi-lingual embeddings are chosen, which contain both Chinese and English words in a multi-lingual embedding matrix.

{\bf Evaluation Metrics}. We follow  \citet{yu2018spider}, evaluating the results using two major types of metrics. The first is exact matching accuracy, namely the percentage of questions that have exactly the same SQL output as its reference. The second is component matching F1, namely the F1 scores for \textsc{select}, \textsc{where}, \textsc{group by}, \textsc{order by} and all keywords, respectively.

{\bf Hyperparameters}. Our hyperparameters are mostly taken from \citet{yu2018syntaxsqlnet}, but tuned on the Chinese Spider development set. We use character and word embeddings from Tencent embedding; both of them are not fine-tuned during model training. Embedding sizes are set to 200 for both characters and words. For the different choices of keywords and column names embeddings, sizes are set to 200 and 300, respectively. Adam \cite{kingma2014adam} is used for optimization, with a learning rate of 1e-4. Dropout is used for the output of LSTM with a rate of 0.5.

For word-based models, segmentation is necessary. We take two segmentors with different performances, including the Jieba segmentor and the model of \citet{yang2017neural}, which we name Jieba and YZ, respectively. To verify their accuracy, we manually segment the first 100 sentences from the test set. Jieba and YZ give F1 scores of 89.8\% and 91.7\%, respectively.

\subsection{Overall Results}
The overall exact matching results are shown in Table \ref{tab:exact}. In this table, ENG represents the results of \citet{yu2018syntaxsqlnet}'s model on their English dataset but under our split. HT and MT denote human translation and machine translation of questions, respectively. Both HT and MT results are evaluated on human translated questions. C-ML and C-S denote the results of our Chinese models based on characters with multi-lingual embeddings and monolingual embeddings, respectively, while WY-ML, WY-S denote the word-based models applying YZ segmentor with multi-lingual embeddings and monolingual embeddings, respectively. Finally, WJ-ML and WJ-S  denote the word model with multi-lingual embeddings and monolingual embeddings with the Jieba segmentor, respectively.

\begin{table}[t]
\centering
\resizebox{0.475\textwidth}{!}{%
\begin{tabular}{ccccccc}
\hline
 &  & Easy & Medium & Hard & Extra Hard & All \\ \hline
\multicolumn{2}{c}{ENG} & 31.8\% & 11.3\% & 9.5\% & 2.7\% & 14.1\% \\ \hline
\multirow{6}{*}{HT} & C-ML & \textbf{27.3\%} & \textbf{9.9\%} & 7.5\% & \textbf{2.3\%} & \textbf{12.1\%} \\
 & C-S & 23.1\% & 7.7\% & 6.2\% & 1.7\% & 9.9\% \\
 & WY-ML & 21.4\% & 8.1\% & \textbf{8.0\%} & 1.7\% & 10.0\% \\
 & WY-S & 20.2\% & 6.4\% & 6.7\% & 2.0\% & 8.9\% \\
 & WJ-ML & 19.8\% & 8.6\% & 5.0\% & 1.3\% & 9.2\% \\
 & WJ-S & 20.1\% & 5.0\% & 5.7\% & 1.7\% & 8.2\% \\ \hline
\multirow{2}{*}{MT} & C-ML & 18.1\% & 4.6\% & 5.2\% & 0.3\% & 7.9\% \\
 & WY-ML & 17.9\% & 4.7\% & 4.5\% & 0.3\% & 7.6\% \\ \hline
\end{tabular}%
}
\caption{Accuracy of Exact Matching on test set.}
\label{tab:exact}
\end{table}

\begin{table*}[t]
\centering
\resizebox{0.8\textwidth}{!}{%
\begin{tabular}{ccccccccc}
\hline
 &  & SEL & SELN & WHE & WHEN & GB & GBN & OB \\ \hline
\multicolumn{2}{c}{ENG} & 47.3\% & 48.2\% & 19.9\% & 24.4\% & 35.0\% & 40.6\% & 57.6\% \\ \hline
HT & C-ML & 40.7\% & 41.2\% & 19.9\% & 23.6\% & 33.6\% & 36.7\% & 53.8\% \\
 & C-S & 40.6\% & 41.0\% & 15.3\% & 17.3\% & 29.2\% & 32.9\% & 51.7\% \\
 & WY-ML & 34.8\% & 35.6\% & 18.1\% & 21.4\% & 26.7\% & 30.9\% & 49.8\% \\
 & WY-S & 34.5\% & 35.6\% & 16.5\% & 19.8\% & 30.2\% & 34.2\% & 46.9\% \\
 & WJ-ML & 34.7\% & 35.4\% & 15.8\% & 19.2\% & 27.9\% & 31.4\% & 52.5\% \\
 & WJ-S & 35.7\% & 36.8\% & 15.9\% & 19.6\% & 24.4\% & 26.8\% & 48.0\% \\ \hline
\multirow{2}{*}{MT} & C-ML & 36.5\% & 37.2\% & 11.3\% & 14.2\% & 29.1\% & 33.4\% & 50.7\% \\
 & WY-ML & 32.1\% & 32.8\% & 11.3\% & 13.4\% & 24.8\% & 27.5\% & 49.1\% \\ \hline
\end{tabular}%
}
\caption{F1 scores of Component Matching on test set.}
\label{tab:all}
\end{table*}

First, compared to the best results of human translation (C-ML and WY-ML), machine translation results show a large disadvantage (e.g. 7.1\% vs 12.1\% using C-ML). We further did a manual inspection of 100 randomly picked machine-translated sentences. Out of the 100 translated sentences, 42 have translation mistakes such as semantic changes (28 sentences) and grammar errors (14 sentences). Both of these facts indicate that data by machine-translation is not reliable for semantic parsing research. 

Second, comparisons among C-ML, WY-ML and WJ-ML, and among C-S, WY-S and WJ-S show that multi-lingual embeddings give superior results compared to monolingual embeddings, which is likely because they bring a better connection between natural language questions and database columns. 

Third, comparisons between WY-ML and WJ-ML, and WY-S and WJ-S indicate that better segmentation accuracy has a significant influence on question parsing. Word-based methods are subject to segmentation errors. 

Moreover, with the current segmentation accuracy of 92\%, a word-based model underperforms a character-based model. Intuitively, since words carry more direct semantic information as compared with database columns and keywords, improved segmentation may allow a word-based model to outperform a character-based model. 

Finally, for easy questions, the character-based model shows strong advantages over the word-based models. However, for medium to extremely hard questions, the trend becomes less obvious, which is likely because the intrinsic semantic complexity overwhelms the encoding differences.

Our best Chinese system gives an overall accuracy of 12.1\%, \footnote{Note that the results are lower than those reported by \citet{yu2018syntaxsqlnet} under their split due to different training/test splits. Our split has less training data and more test instances in the ``Hard'' category and less in ``Easy'' and ``Medium''.} which is less but comparable to the English results. This shows that Chinese semantic parsing may not be significantly more challenging compared to English with text to SQL.

\begin{figure}[t]
	\centering 
	\includegraphics[width=1\linewidth]{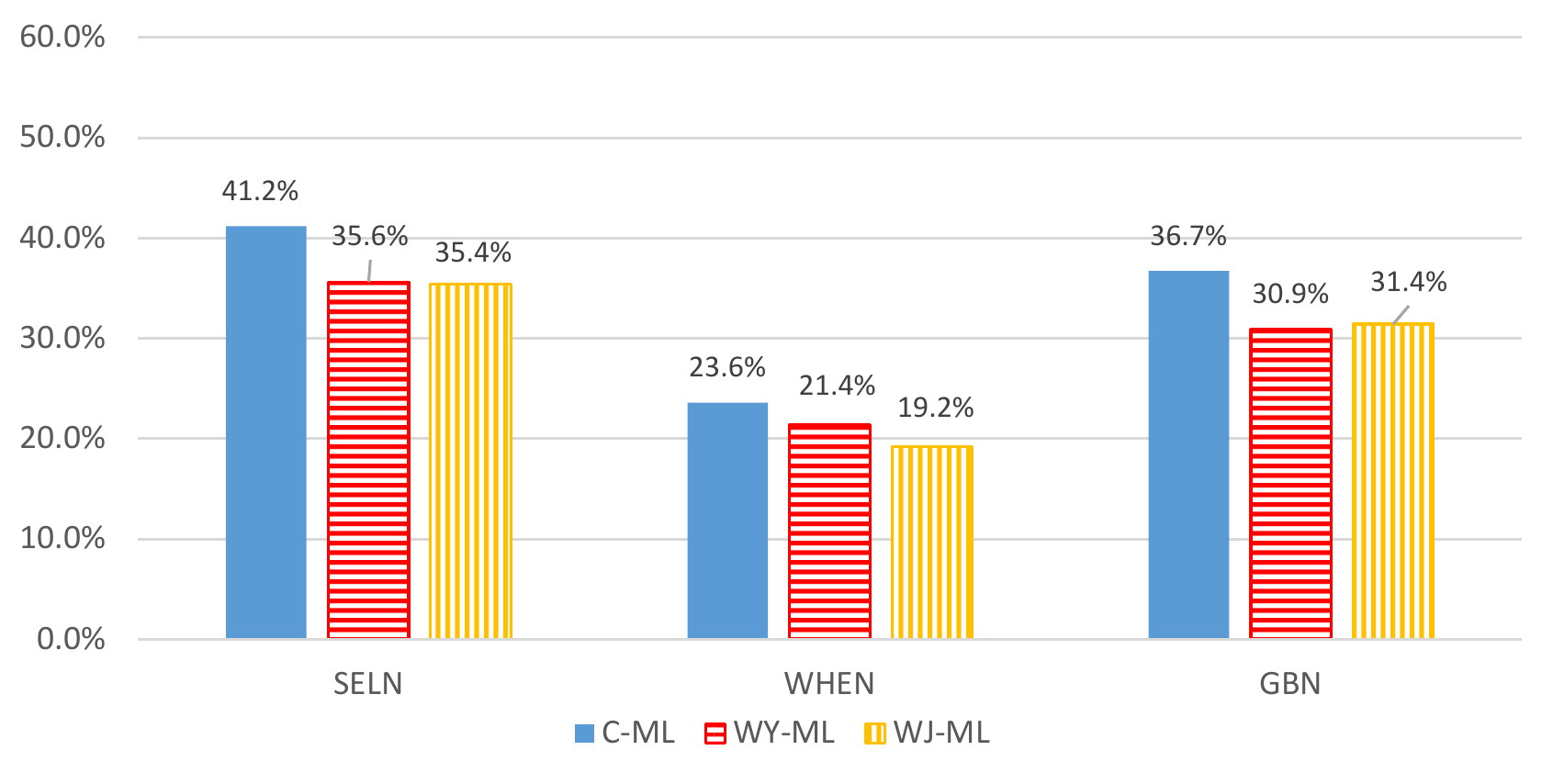}
	\caption{Component Matching Comparisons.}
	\label{fig:dev}
\end{figure}

{\bf Component matching}. Figure \ref{fig:dev} shows F1 scores of several typical components, including  \textsc{seln} (SELECT NO AGGREGATOR), \textsc{when} (WHERE NO OPERATOR) and \textsc{gbn} (GROUP BY NO HAVING), applying the superior multi-lingual embeddings. The trends are consistent with the overall results. 

The detailed results are shown in Table \ref{tab:all}. Specifically, the char-based methods achieve around 41\% on SELN and SEL (SELECT), which are about 5\% higher compared to the word-based methods. This result may be due to the fact that word-based models are sensitive to the OOV words \cite{zhang2018chinese, li2019word}. Unlike other components, SEL and SELN are confronted with more severe OOV challenges caused by recognizing the unseen schema during testing. 

In addition, the models using multi-lingual embedding overperform the models using separate embeddings on both WHEN and OB (ORDERBY), which further demonstrates that embeddings in the same dimension distribution benefit to strengthen the connection between the question and the schema. 

Contrary to the overall result, the models employing the jieba segmentor perform better than those using the YZ segmentor on OB. The reason is that the jieba segmentor has different word segmentation results in terms of the superlative of adjectives. For example, the word “最高” (the highest) is segmented as “最”(most) and “高”(high) by YZ segmentor but “最高” in jieba segmentor. This again demonstrates the influence of word segmentation. Finally, for GB (GROUPBY) there is not a regular contrast pattern between different models, which can be likely because of the lack of sufficient training data.

\subsection{Case study}
\begin{figure}[t]
	\centering 
	\includegraphics[width=1\linewidth]{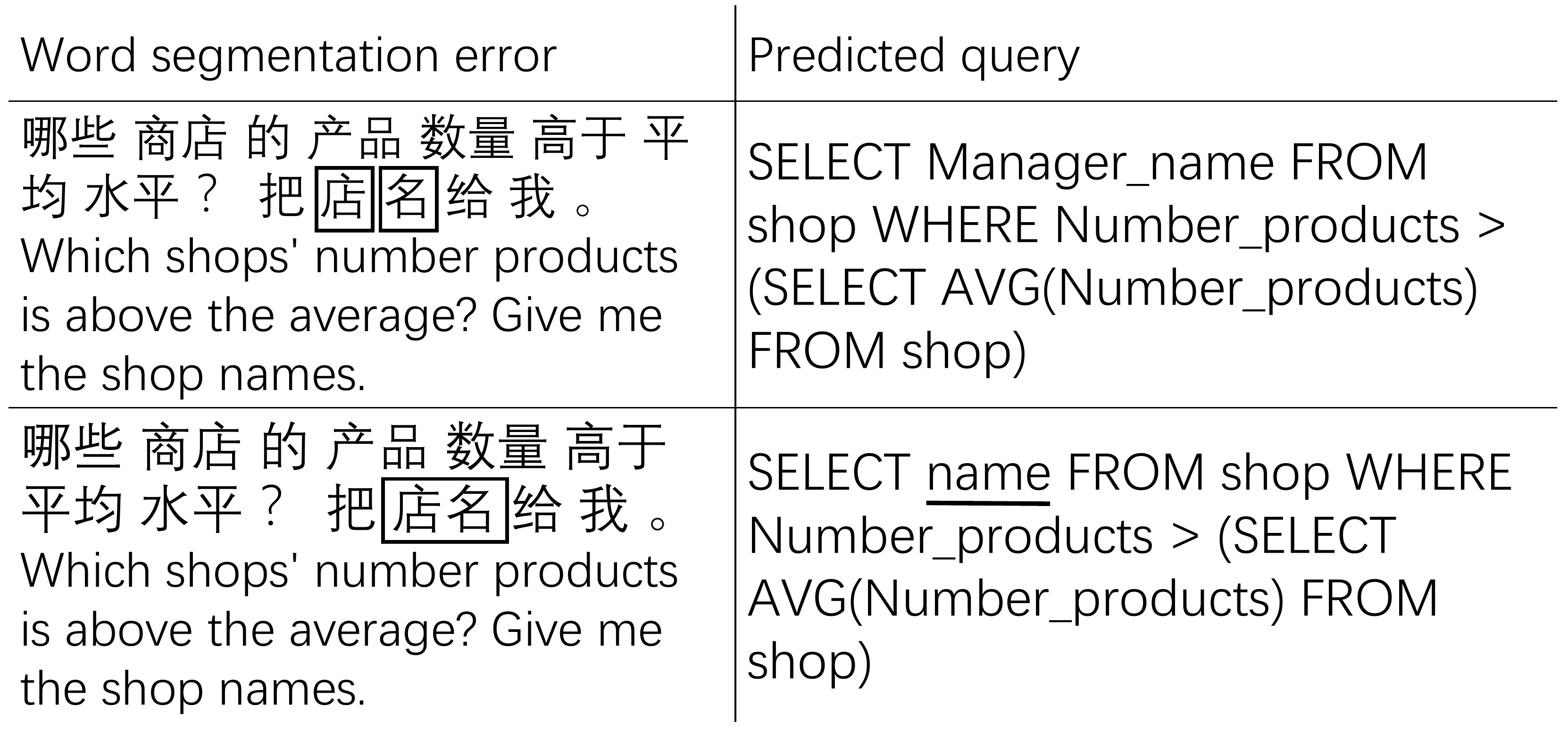}
	\caption{Word segmentation error.}  
	\label{fig:segmentation}  
\end{figure}

Figure \ref{fig:segmentation} shows the negative influence of segmentation errors. In particular, the incorrect segmentation of the word ``店名'' (shop name) leads to incorrect SQL for the whole sentence, since the character ``店'' (shop) can typically be associated with ``店长'' (shop manager).

\begin{figure}[t]
	\centering 
	\includegraphics[width=1\linewidth]{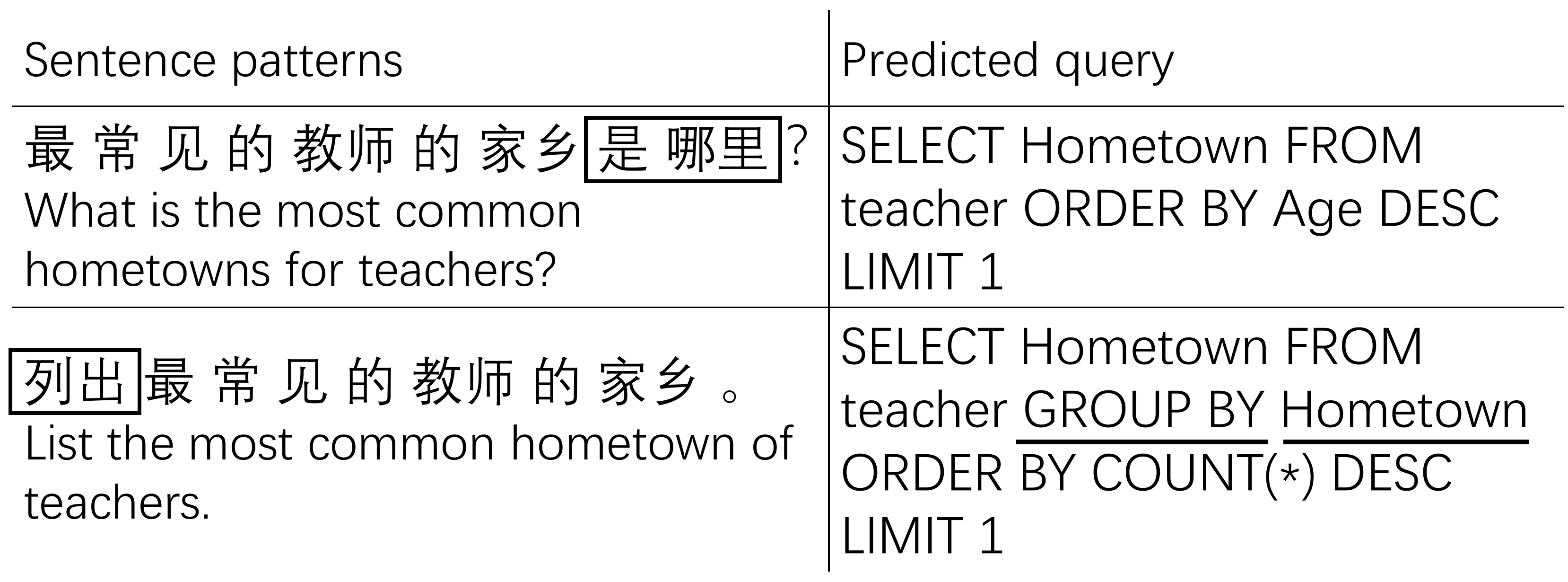}
	\caption{Sentence pattern.}
	\label{fig:pattern}
\end{figure}

Figure \ref{fig:pattern} shows the sensitivity of our model to sentence patterns. In particular, the word-based model gives incorrect predictions for many question sentences frequently. As shown in the first row, the word ``where'' confuses the system for making a choice between ``ORDER BY'' and ``GROUP BY''. When we manually change the sentence pattern into ``List the most common hometown of teachers'', the parser gives the correct keyword. In contrast, the character-based model is less sensitive to question sentences, which is likely because characters are less sparse compared with words. More training data or contextualized embeddings may alleviate the issue for the word-based method, which we leave for future work.

\begin{figure}[t]
	\centering 
	\includegraphics[width=1\linewidth]{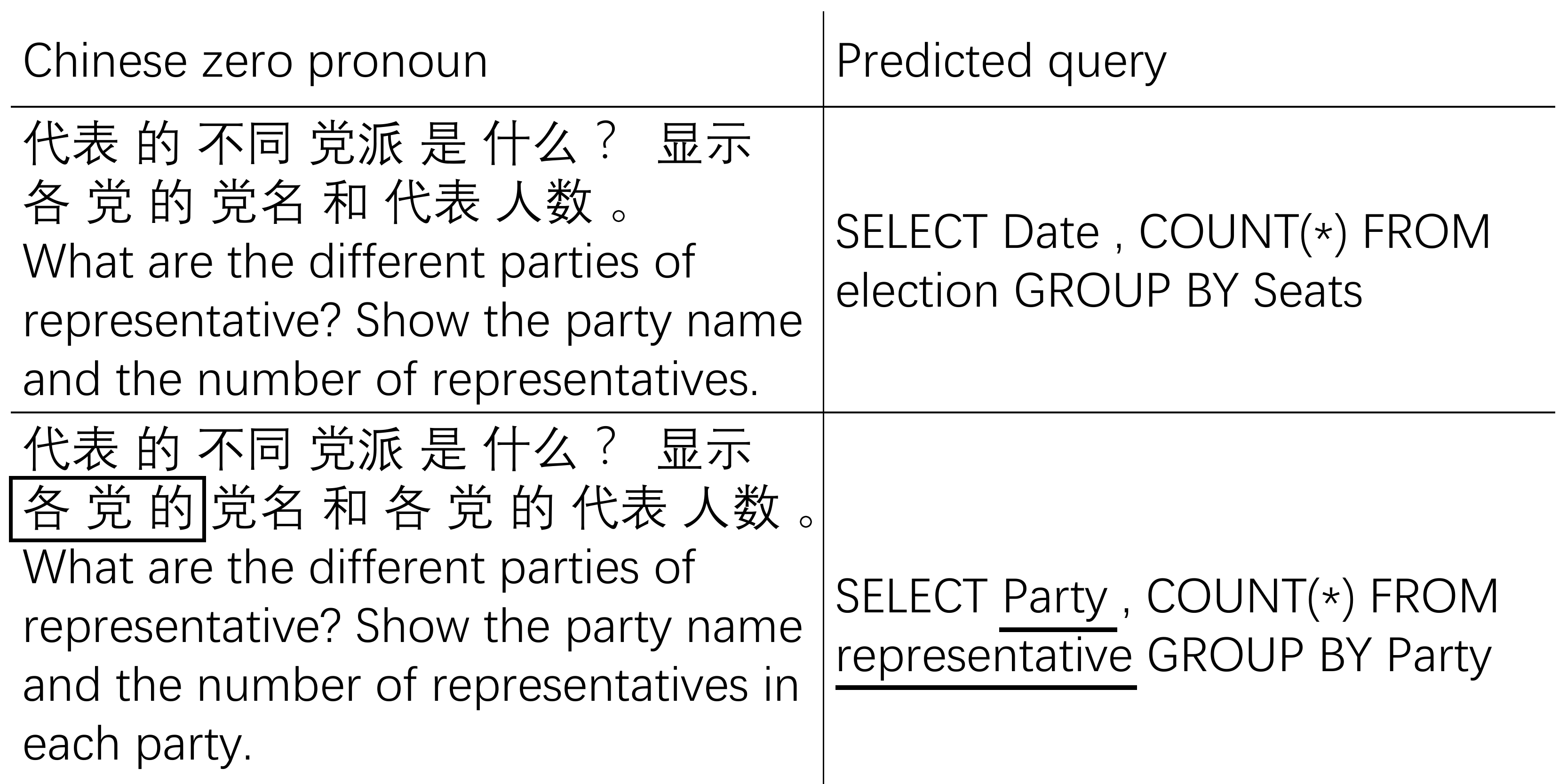}
	\caption{Chinese zero pronoun.}
	\label{fig:linguistic}
\end{figure}

Figure \ref{fig:linguistic} shows the sensitivity of the model to Chinese linguistic patterns. In particular, the first sentence has a zero pronoun ``各党的'' (in each party), which is omitted later. As a result, a semantic parser cannot tell the correct database columns from the sentence. We manually add the correct entity for the zero pronoun, resulting in the second sentence. The parser can correctly identify both the column name and the table name for this corrected sentence. Since zero-pronouns are frequent for Chinese \cite{chen2016chinese}, they give added difficulty for its semantic parsing.

\section{Conclusion}
We constructed a first resource named CSpider for Chinese sentence to SQL, evaluating the performance of a strong English model on this dataset. Results show that the input representation, embedding forms and linguistic factors all have the influence on the Chinese-specific task. Our dataset can serve as a starting point for further research on this task, which can be beneficial to the investigation of Chinese QA and dialogue models.

\section*{Acknowledgments}
We thank the anonymous reviewers for their detailed and constructive comments. Yue Zhang is the corresponding author.

\bibliography{emnlp-ijcnlp-2019}
\bibliographystyle{acl_natbib}

\end{CJK}

\end{document}